\documentclass[letterpaper]{article} 
\usepackage[submission]{aaai24}  
\usepackage{times}  
\usepackage{helvet}  
\usepackage{courier}  
\usepackage[hyphens]{url}  
\usepackage{graphicx} 
\usepackage{natbib}  
\usepackage{caption} 
\usepackage{algorithm}
\usepackage{algorithmic}
\usepackage{amsmath}
\usepackage{amsfonts}

\usepackage{newfloat}
\usepackage{listings}
\DeclareCaptionStyle{ruled}{labelfont=normalfont,labelsep=colon,strut=off} 
\floatstyle{ruled}
\newfloat{listing}{tb}{lst}{}
\floatname{listing}{Listing}
\title{Open-Source Fermionic Neural Networks with Ionic Charge Initialization}
\author{
    Shai Pranesh\textsuperscript{\rm 1},
    Shang Zhu\textsuperscript{\rm 2},
    Venkat Viswanathan\textsuperscript{\rm 3},
    Bharath Ramsundar\textsuperscript{\rm 1}
}
\affiliations{
  \textsuperscript{\rm 1}Deep Forest Sciences, California, USA \\
  \textsuperscript{\rm 2}Department of Mechanical Engineering, University of Michigan, Michigan,USA \\
  \textsuperscript{\rm 3}Department of Aerospace Engineering, University of Michigan, Michigan,USA \\
  shai@deepforestsci.com, shangzhu@umich.edu, venkvis@umich.edu, bharath@deepforestsci.com
}

\usepackage{bibentry}

\begin{document}

\maketitle

\begin{abstract}
Finding accurate solutions to the electronic Schrödinger equation plays an important role in discovering important molecular and material energies and characteristics. Consequently, solving systems with large numbers of electrons has become increasingly important. Variational Monte Carlo (VMC) methods, especially those approximated through deep neural networks, are promising in this regard. In this paper, we aim to integrate one such model called the FermiNet, a post-Hartree-Fock (HF) Deep Neural Network (DNN) model, into a standard and widely used open source library, DeepChem. We also propose novel initialization techniques to overcome the difficulties associated with the assignment of excess or lack of electrons for ions.
\end{abstract}

\section{Introduction}

The Variational Monte Carlo (VMC) technique is based on the variational 
principle in quantum mechanics, which states that the expected value of the
energy of a trial wave function is always greater than or equal to the ground 
state energy of the molecule system. The expected energy is the average energy 
of the sampled electrons.

\begin{align}
\textbf{E}_{\text{ground}} \leq \frac{\langle \psi \mid \hat{H} \mid \psi \rangle}{\langle \psi \mid \psi \rangle} \approx \textbf{E}_{\text{expected}} = \frac{1}{N} \sum E(\theta)
\end{align}
\noindent where the energy is calculated by the equation
\begin{align}
\mathbf{E}(\theta) = \mathbf{E}_{\text{kinetic}} + \mathbf{E}_{\text{el-el}} + \mathbf{E}_{\text{nuc-nuc}} + \mathbf{E}_{\text{nuc-el}}
\end{align}
\noindent and where kinetic energy is calculated as follows:
\begin{align}
\mathbf{E}_{\text{kinetic}} = -\frac{1}{2}\sum_{i}\left [ \left(\frac{\partial \log(|\psi_\theta|)}{\partial r_i}\right)^2 + \frac{\partial^2 \log(|\psi_\theta|)}{\partial x^2} \right ]
\end{align}
Here \(r_i\) is the electron coordinates in the \(i\)-th domain $(x, y, z)$ coordinates, and \(\psi_\theta\) is the trial wavefunction depending on parameters $\theta$. \(\mathbf{E}_{\text{nuc-el}}\), \(\mathbf{E}_{\text{nuc-nuc}}\), \(\mathbf{E}_{\text{el-el}}\) correspond to the total potential energy between each nucleon with each electron, each nucleon with other nucleons, each electron with other electrons in the molecule system, respectively.

Traditional VMC techniques initialize and sample the electron’s 
coordinates via the Monte Carlo algorithm and move toward values that 
minimize the expected energy. The general steps in a VMC algorithm are as follows:
\begin{itemize}
    \item Initialize suitable parameters $\theta$ in a chosen trial wavefunction $\psi_\theta$ and initialize electron coordinates.
    \item Compute new trial coordinates of the electrons using a suitable distribution.
    \item Employ the Metropolis algorithm to determine whether to accept or reject the proposed new move, using the square of the magnitude of the wavefunction as the electron's probability.
    \item If the step is accepted, update parameters after a set number of steps.
    \item Conclude the calculation and compute the final averages for expected energy calculation.
\end{itemize}

The results heavily depend on the quality of the trial wavefunction $\psi_\theta$. In this 
regard, DNN-based approaches can help to approximate and optimize the trial wave 
functions while running the VMC simultaneously. Certain physics-based elements
are built into the ansatz for better convergence, with the training cost as to 
minimize the expected energy function as low as possible. \\

\subsection{DeepChem and Differentiable Physics}
DeepChem \cite{ramsundar2021deepchem} is an open source Python library for scientific machine learning and deep learning on molecular and quantum datasets. DeepChem offers a framework for tackling challenging scientific problems in domains such as drug discovery, energy, and biotechnology. This is achieved by standard workflows constructed from fundamental components, including data loaders, featurizers, data splitters, learned models, metrics, and hyperparameter tuners. This systematic design has allowed DeepChem to be part of a wide variety of applications, like large-scale benchmarking for molecular machine learning through the
MoleculeNet benchmark suite \cite{wu2018moleculenet}, protein-ligand interaction modeling \cite{gomes2017atomic}, generative modeling of molecules \cite{frey2023fastflows}, and more.\\

DeepChem aims to support open-source differentiable physics \cite{ramsundar2021differentiable} infrastructure for scientific machine learning. Differentiable density functional theory infrastructure has been previously integrated into DeepChem for the same purpose \cite{vidhyadhiraja2023open}. This work integrates FermiNet \cite{pfau2020ferminet} as part of the ongoing efforts to facilitate differentiable physics support and enable rapid physical calculations via neural network approximations.

\section{Implementation}
The FermiNet model was split and implemented as the following major components, each dealing with a specific task:
\begin{itemize}
\item Electron Sampler
\item Hartree-Fock Baselines
\item Ionic Charge Initialization 
\item FermiNet Model (neural network layers)
\end{itemize}

\subsection{Electron Sampler}
The electron sampler was implemented into the DeepChem utils subpackage using Numpy. The electron sampler can initialize electron positions centered on each nucleon. The number of electrons initialized depends on the atomic number of each nucleon, and in the case of ions, electrons are either added or deleted according to Mulliken population analysis (discussed briefly in the subsequent subsection).

After initialization, the trial electron coordinates are sampled using Gaussian distribution centered around the present electron's coordinates, which are then accepted/rejected via the Metropolis-Hastings algorithm using the log probability of the wavefunction ($2 \cdot \log(|\psi|)$) calculated at the electron's position. The difference between the log probability of trial and original electrons gives the probability of the jump or acceptance probability($A$). A random value \(u \sim \log(\mathrm{Uniform}(0, 1)\)) is then generated, and if \(u \leq A\), the trial coordinates are accepted (\(r_{t+1} = r'\)); otherwise, it is rejected, and the Markov chain stays at the current state (\(r_{t+1} = r_t\)). This process is repeatedly done for a set number of iterations, usually until the model gives converged results.

\subsection{Hartree-Fock Baselines}
At each epoch of the initial pretraining for the FermiNet model, the HF baseline is computed using the STO-6G basis function. This calculation is performed using the sampled electron's postion for that particular epoch, aiming to align the models's calculated orbitals with the HF baseline. This aids in accelerating convergence during the actual training of the model layers. The conventional HF calculation ignores the electron-correlation term, which will be integrated into the model training process itself. These calculations were performed using the PySCF library \cite{pyscf}.

\subsection{Ionic Charge Initialization}
In an ion, one or more of the nucleons has either an excess or lack of electrons to accommodate the ionic charge. There is no present initialization policy of ionic charges in DeepMind's implementation of the model \cite{ferminet_github}. We propose a novel initialization policy based on Mulliken population analysis \cite{Mulliken_1955}.

The Mulliken population charge matrix is calculated during the HF calculations, which gives the partial electron charges of each atom in a molecule based on the overlap of atomic orbitals. This charge matrix can act as a heuristic to add/remove electrons as per the ionic charge of the molecule system. The nucleon with the most negative partial charge is the most ideal candidate for excess electrons and the one with the most positive partial charge will have a lack of electrons.

The exact algorithm for the charge initialization policy can be seen in Algorithm 1. The input for this algorithm will be a matrix of partial charges computed via Mulliken Population analysis. For example, in a LiH+ ion, the partial charge matrix will be [0.88694, 0.11306], where the first term corresponds to the Li partial charge and the other term to the partial charge of H in the ionic system. We also use another matrix for the calculation called the electron matrix which consists of the electrons of each nucleon in an uncharged system. For example in LiH+, the electron matrix would be [3, 1] as Li has 3 electrons and H 1 electron in its uncharged form.

\begin{algorithm}[tb]
\caption{Charge Initialization Policy}
\label{alg:algorithm}
\textbf{Input}: Partial charge matrix from Mulliken population analysis.\\
\textbf{Output}: Electron number for each nucleon in the ion
\begin{algorithmic}[1] 
\STATE Let $charge$ be the partial charge matrix.
\STATE Let $electron$ be the array initialized with the atomic number of each nucleon.
\STATE Let $t=$ ionic charge of the system.
\WHILE{\( t \neq 0 \)}
\IF {$t < 0$}
\STATE $index = \text{argmin}(\textit{charge})$
\STATE $electron[index]$ += $1$
\STATE $t$ += $1$
\STATE $charge[index]$ += $1$
\ELSE
\STATE $index = \text{argmax}(\textit{charge})$
\STATE $electron[index]$ -= $1$
\STATE $t$ -= $1$
\STATE $charge[index]$ -= $1$
\ENDIF
\ENDWHILE
\STATE \textbf{return} electron
\end{algorithmic}
\end{algorithm}

\subsection{FermiNet Model}
The FermiNet architecture consists of two major feature layers: $1$-electron and $2$-electron feature layers which act as the embedding layer for electron distance features to be used to model the wavefunction ansatz. The 2-electron layer featurizes each of the electron's distance from the other electrons whereas the 1-electron layer featurizes each electron's distance from each of the nucleons, using the value of 2-electron features in the process.

The ansatz consists of an “envelope layer” which uses the 1-electron features to enforce the boundary condition of the probability of the electron's position tending to 0 when the electron moves towards infinity from the nucleus. The envelope layer of each electron $j$ is modeled as:

\begin{align}
\sum_{m}^{} \pi_{k,\alpha}^{i,m} . \exp(-\sigma_{k,\alpha}^{i,m} \lvert \text{R}_{j,m} \rvert)
\end{align}

\noindent where \(\sigma\) is a learnable \(3 \times 3\) tensor parameter and \(\pi\) is a learnable scalar parameter. There are a total of $ikm$\(\alpha\) copies of the \(\sigma\) and \(\pi\) parameters. $m$, $k$, $i$, $j$, and \(\alpha\) correspond respectively to the number of nucleons, the number of determinants to be used in the ansatz, the number of orbitals, the number of electrons, and the spin of the electron taken. The term \(\text{R}_{j,m}\) corresponds to the 1-electron-distance vector of the \(j_{}^{th}\) electron and \(m_{}^{th}\) nucleon.\\ \\
The envelope layer along with the computed 1-electron feature is used to model the ansatz's orbital evaluated at a specific electron. Using the slater-type determinants of these orbitals, the wavefunction's value can be computed at the input electron's position.

\subsubsection{FermiNet Training}
FermiNet has 2 parts to the training: supervised pretraining and unsupervised training.  
During pretraining, orbital values calculated from the Hartree-Fock method are used as labels, with FermiNet’s orbital values trained to match the Hartree-Fock baseline. The loss function is:

\begin{align}
\mathcal{L}_{pretrain} &= \frac{1}{N} \sum_{kij}^{} \left( \Phi_{kij} - {\Phi_{kij}^{HF   }} \right)^{2}
\end{align}

\noindent where \(\Phi\) and \({\Phi_{}^{HF}}\) correspond to the calculated orbital values from the FermiNet model and the HF baseline respectively. $k$, $i$, and $j$ correspond to the number of determinants, number of orbitals, and number of electrons respectively.

Next in the training phase, the model's parameters are tuned according to the modified gradients using the expected energy. The parameters in the network are updated according to the following gradient
\begin{align}
\nabla_{\theta}\mathcal{L}_{train} &= 2\mathbb{E}_p \left[ \left( E_p - \mathbb{E}_p(E_p) \right)\nabla_{\theta}\log |\psi| \right]
\end{align}
\noindent where $\mathbb{E}_p$ is the expected mean over the sampled electrons, $E_p$ is the energy of sample $p$ across all batches, and \(\nabla_{\theta}\mathcal{L}_{train}\) is the modified gradient with respect to the model parameters \(\theta\), while \(\nabla_{\theta}\log |\psi|\) is the gradients with respect to the \(\log |\psi|\) term.

The gradients of the network are updated using the PyTorch hooks module. For the kinetic energy calculation, functorch \cite{functorch2021} (available from Pytorch 2.0.0) is used to calculate the Hessian and the Jacobian. 

\section{Results}
In Fig 1, we plot the ground state energies for Hydrogen molecules with different internuclear distances and compare them against the CCSD values calculated by PySCF. The FermiNet's ground state energy curve closely tracks the CCSD curve.
\begin{figure}[t]
\centering
\includegraphics[width=0.9\columnwidth]{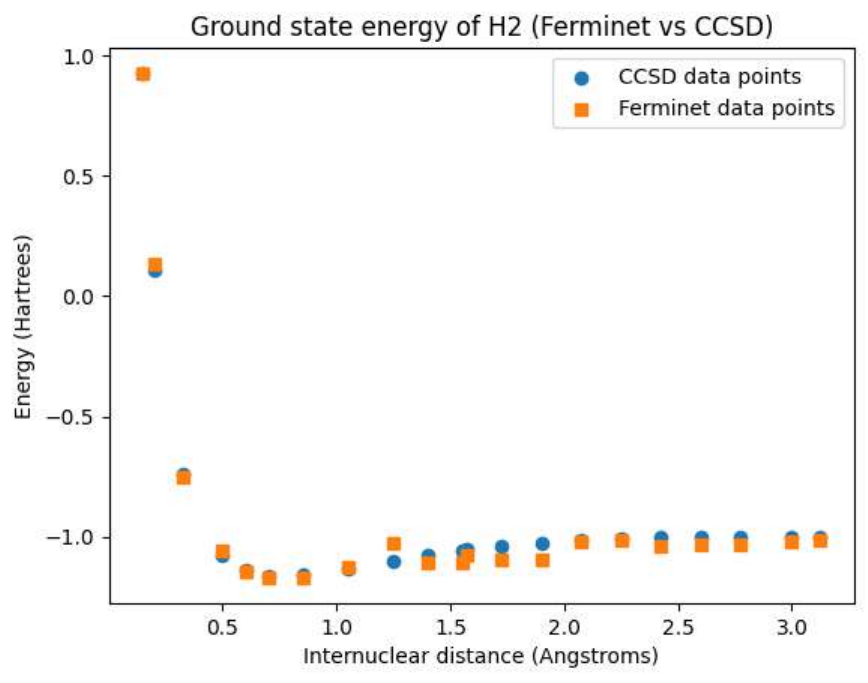} 
\caption{H2 ground state energy - FermiNet vs CCSD. Plotted using 21 different data points. The values obtained from FermiNet can be seen to closely match that of the values calculated via CCSD. Running the model with more MCMC steps and iterations can give more reliable results.}
\label{fig3}
\end{figure}
We also tested the model to estimate the LiH molecule's ionization potential, and compared them with the CCSD and HF calculation values obtained from the NIST database \cite{nist-cccbdb}. The result can be found in Table 1.
\begin{table}[ht]
\centering
\begin{tabular}{|c|c|c|}
\hline
\textbf{FermiNet} & \textbf{CCSD} & \textbf{HF} \\
\hline
0.2481 & 0.2637 & 0.2387 \\
\hline
\end{tabular}
\caption{The ionization potential for LiH in hartrees obtained from different methods, as stated, is given. The FermiNet's ionization potential can be seen to have improved upon the values obtained from Hartree-Fock calculations and is closer to the energy obtained using CCSD. This ionization potential was calculated using the ionic charge initialization policy that we proposed (and is not present in DeepMind's implementation.)}
\label{tab:example}
\end{table}
We used a batch size of 8 and 10 steps between MCMC updates to get the results. Using a larger batch size, more iterations, and more MCMC steps will decrease the noise in the results at the cost of increased training time. 

\section{Conclusion}
From Figure 1 and Table 1, we observe that the model's results closely match that of the CCSD method and improve on top of the HF baseline for LiH ionization potential calculation. We anticipate integrating FermiNet into DeepChem will lead to improved infrastructure for rapid experimentation with the FermiNet model for calculating accurate ground state energies of molecule/ion systems.

We aim to further improve the model by using Pytorch's JIT functionality to match the capabilities of DeepMind's JAX implementation \cite{spencer2020better}. Also, we aim to integrate features from PauliNet\cite{10.1063/5.0157512}, another DNN-based VMC model. PauliNet has the advantage of employing fewer parameters and achieving quicker training compared to FermiNet but with slightly lower accuracy. This is because PauliNet has ``physical conditions" built into the ansatz whereas FermiNet aims to train and learn those features. Thus, incorporating physical priors into FermiNet can lead to faster convergence of the model with accurate results. The standardized DeepChem implementation we have introduced will help facilitate such experimentation in future work.

\appendix
\bibliography{ref}

\end{document}